\documentclass[runningheads]{llncs}

\usepackage{eccv}

\usepackage{eccvabbrv}

\usepackage{graphicx}
\usepackage{booktabs}

\usepackage[accsupp]{axessibility}  

\usepackage{amsmath}
\usepackage{amssymb}
\usepackage{bbm}
\usepackage{booktabs}
\usepackage{multirow}
\usepackage{multicol}
\usepackage{stfloats}
\usepackage[normalem]{ulem}
\usepackage{tabularray}
\usepackage{rotating}
\usepackage{tabularx}

\usepackage{hyperref}

\usepackage{orcidlink}

\footnotetext[2]{$^\dag$ Corresponding authors.}

\begin{document}

\title{Learn to Memorize and to Forget: A Continual Learning Perspective of Dynamic SLAM} 

\titlerunning{Learn to Memorize and to Forget}

\author{Baicheng Li\inst{1} \and
Zike Yan\inst{2}$^\dag$ \and
Dong Wu\inst{1} \and
Hanqing Jiang\inst{3} \and
Hongbin Zha\inst{1}$^\dag$ }

\authorrunning{B. Li et al.}

\institute{National Key Lab of GAI, School of IST \\
PKU-SenseTime Machine Vision Joint Lab \\
Peking University\and
AIR, Tsinghua University \and
SenseTime Research \\
}

\maketitle

\begin{abstract}
Simultaneous localization and mapping (SLAM) with implicit neural representations has received extensive attention due to the expressive representation power and the innovative paradigm of continual learning. However, deploying such a system within a dynamic environment has not been well-studied. Such challenges are intractable even for conventional algorithms since observations from different views with dynamic objects involved break the geometric and photometric consistency, whereas the consistency lays the foundation for joint optimizing the camera pose and the map parameters. In this paper, we best exploit the characteristics of continual learning and propose a novel SLAM framework for dynamic environments. While past efforts have been made to avoid catastrophic forgetting by exploiting an experience replay strategy, we view forgetting as a desirable characteristic. By adaptively controlling the replayed buffer, the ambiguity caused by moving objects can be easily alleviated through forgetting. We restrain the replay of the dynamic objects by introducing a continually-learned classifier for dynamic object identification. The iterative optimization of the neural map and the classifier notably improves the robustness of the SLAM system under a dynamic environment. Experiments on challenging datasets verify the effectiveness of the proposed framework.
\end{abstract}    
\section{Introduction}
\label{sec:intro}
\begin{figure}[t]
    \centering
    \includegraphics[width=0.95\textwidth]{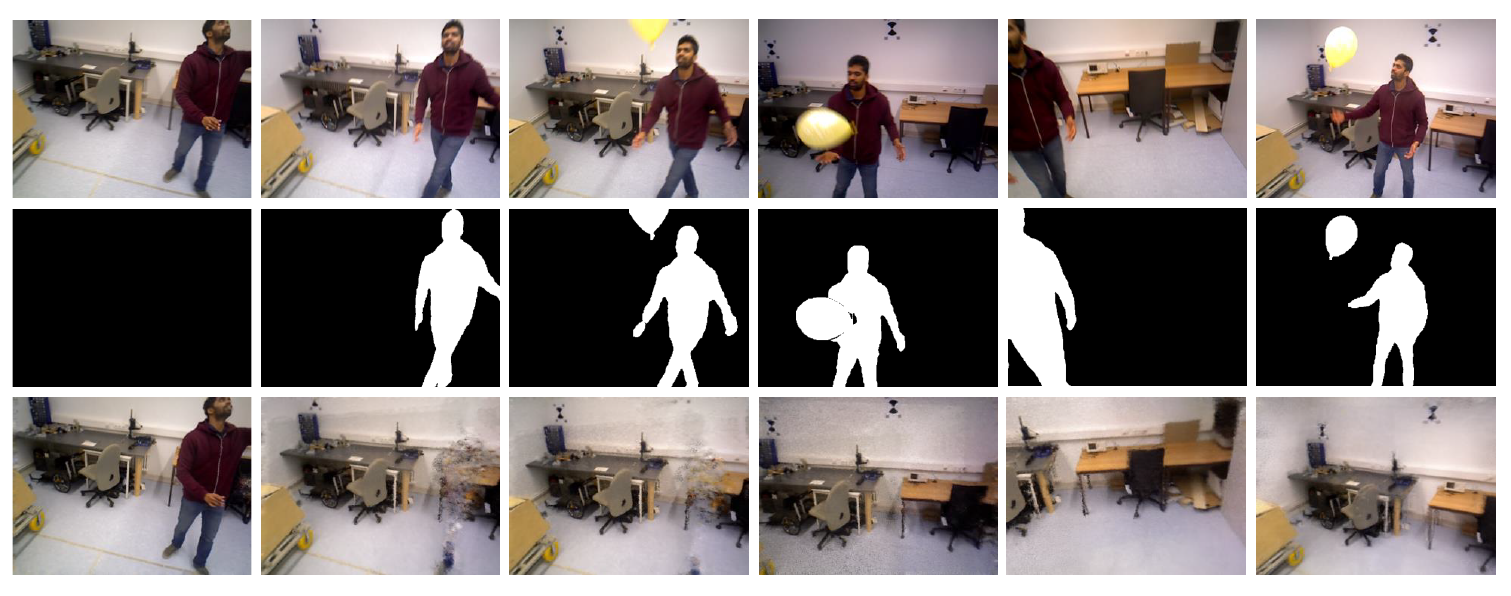}
    \caption{We introduce a continual learning based SLAM framework under challenging dynamic environments (top row). The proposed method jointly learns a classifier to alleviate the effects induced by the moving objects (middle row), and a neural map to memorize past observations as a neural radiance field (bottom row). The iterative optimization of pose, map, and classifier parameters forms a robust SLAM system that learns to memorize and to forget adaptively in the changing open world.}
    \label{fig:cover}
\end{figure}
Simultaneous localization and mapping (SLAM) describes 
the instant agent state and the environment where the agent operates. By constructing a consistent map on the fly given sequential observations, the agent gradually gains knowledge of the environment that can be utilized for downstream vision and robotics applications. The consistency in both temporal and spatial domains lays the foundation of the problem since the genesis of SLAM~\cite{Durrant2006survey}, where the view-invariant photometric and geometric cues within the map are leveraged to predict and validate future measurements~\cite{Cadena2016tro}. Nevertheless, this consistency cannot be guaranteed in the setting as various environmental changes may occur due to the object movements. Alleviating the effects induced by the inconsistency between observations and the map is of great importance for robust long-term deployment. 

Intuitively, a robust SLAM system can be achieved if frame-to-model alignment is conducted based on pure static/invariant features~\cite{Saputra2018CSUR}. Such a system requires accurate identification of environmental changes in both observations and the map. Conventional methods turn to remove the dynamic objects in observations through motion segmentation, and update the discretized map by heuristically deleting the corresponding areas. Recent advances~\cite{Yan2021iccv, sucar2021imap} show that an implicit neural representation can also be updated in a purely static environment through test-time optimization to serve as the map of a dense SLAM system. Besides the experience-replay based continual learning paradigm that most methods adopt to avoid catastrophic forgetting of past observations, we argue that the forgetting is also a nice property to update the neural map given environmental changes. Only the retention of invariant features should be made from the observations, whereas the changing part will be naturally forgotten under constant distribution shifts.


In this work, we introduce a dense neural SLAM framework to tackle challenging dynamic scenarios. The key idea is the continual learning of two modules, a neural map $f(\mathbf{x};\theta^t_M)$ that distills past observations into a continuous neural radiance field, and a binary classifier $g(\mathbf{z};\theta^t_C)$ that records the motion status (static/dynamic) of each instance given the encoded feature $\mathbf{z}$. The continual learning fashion guarantees online adaption to the instant state of scene geometry, appearance, and object motion status. Both modules accumulate knowledge from sequential observations and automatically decide what to memorize and what to forget. The inconsistent areas between the observation and the map will be identified and not contribute to the pose estimation and map updating. Such iterative optimization of poses, map parameters, and object motion status leads to a robust framework for dynamic SLAM under changing environments. To summarize, our main contributions include:
\begin{itemize}
\item We present for the first time the deployment of a dense neural SLAM framework under challenging dynamic environments. The proposed method leads to reliable motion segmentation, robust camera tracking, and convenient map updating under diverse environmental changes.
\item We propose a continual learning method for updating a classifier that records the motion status of objects within the environment. The instance-aware classifier applies to open-world scenarios and shows positive forward and backward transfer. The module can also integrate prior knowledge regarding potentially movable instances through pre-training.
\item We show that the forgetting mechanism of continual learning can be exploited to update the neural scene representation under changing environmental conditions.
\end{itemize}
\section{Related Work}
\label{sec:related_work}

\begin{figure}[t]
    \centering
    \includegraphics[width=\textwidth]{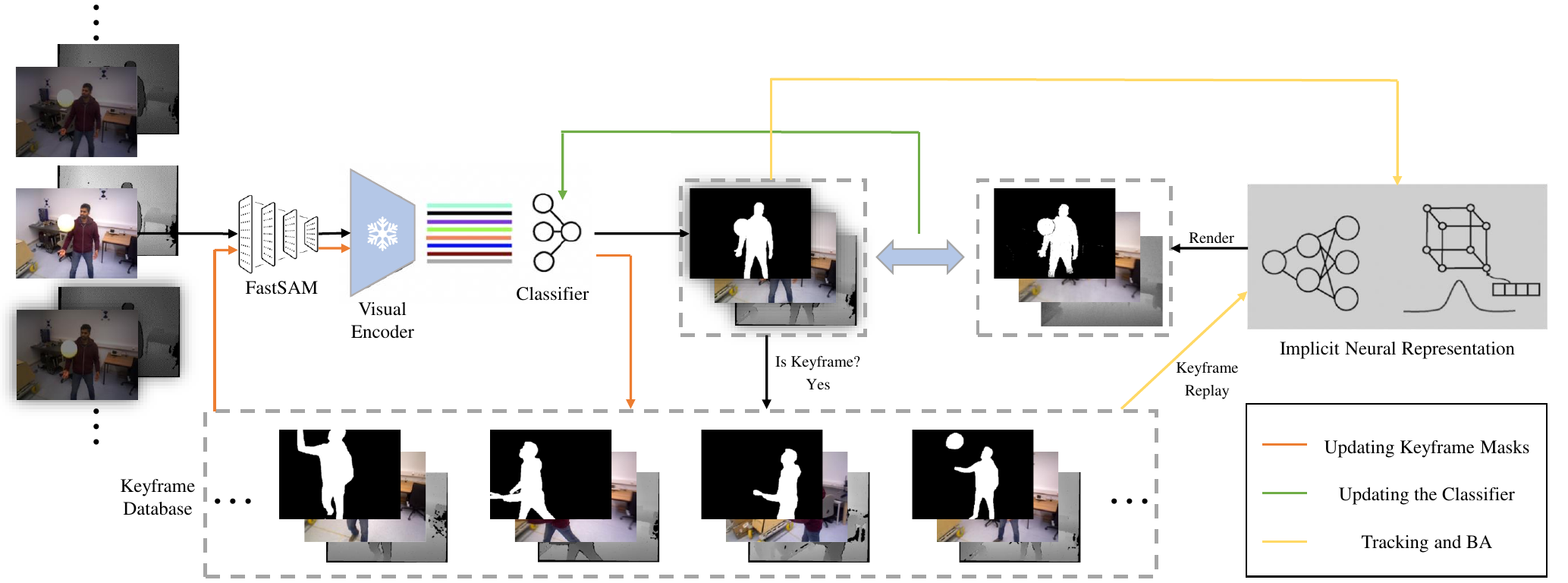}
    \caption{Overview of the proposed method. We effectively integrated instance segmentation module, visual encoder, and a continually-learned classifier to achieve accurate dynamic object identification, enabling robust localization and mapping in complex dynamic environments.}
    \label{fig:classifier}
\end{figure}

\noindent{\bf Visual SLAM in dynamic environments.} This line of work removes dynamic objects and reconstructs the static environment. A number of methods~\cite{cheng2019improving, scona2018staticfusion, palazzolo2019refusion} utilize warping or reprojection to identify the inconsistency in motion, appearance, or geometry. Although motion segmentation can be generalized to different environments, motion ambiguity leads to typical failures when a large portion of a moving object occupies the image. There is also an attempt~\cite{kaneko2018mask} that relies on semantic segmentation ~\cite{he2017mask} to pre-define dynamic categories. However, the solution mainly relies on common sense and cannot adapt to the real open world. In addition, some approaches combine motion detection with semantic cues. DynaSLAM~\cite{bescos2018dynaslam} and DRG-SLAM~\cite{wang2022drg} filter out features that fall into a pre-defined category or avoid geometric constraints. They might over-segment dynamic areas and leave insufficient information for localization. On the contrary, SLAMANTIC~\cite{schorghuber2019slamantic} and CFP-SLAM~\cite{hu2022cfp} check the observations in the pre-defined category through projection and only remove the features that present inconsistency. Nonetheless, the temporal consistency of motion and its applicability to the open world are commonly ignored. The relevant methods struggle to trade-off between over-segmentation and under-segmentation. Additionally, there exists a category of work dedicated to change detection~\cite{adam2022objects,palazzolo2017change,palazzolo2018fast,ulusoy2014image,taneja2011image,taneja2013city}, which can identify areas of change in the environment offline, given known camera poses and a complete point cloud. However, in the context of SLAM, both the camera poses and the complete environmental model are unknown, and the process needs to be conducted online. We effectively fulfill this requirement through a continually-learned classifier.

\noindent{\bf Dense SLAM with implicit neural representations.}
The compact and continuous representation power of implicit neural representations~\cite{mildenhall2020nerf}  draws public attention in the SLAM community. The seminal works of~\cite{sucar2021imap} and~\cite{Yan2021iccv} show that neural representation can be updated through test-time optimization. The follow-ups try different neural representations for better efficiency or accuracy~\cite{zhu2022nice,johari2022eslam, kruzhkov2022meslam, yang2022vox, chung2022orbeez,ming2022idf, ortiz2022isdf, sandstrom2023point}. Recently, Co-SLAM~\cite{wang2023co} leverages both the high-frequency preserving characteristics of coordinate-based representation and the fast convergence of optimizable feature grids for an accurate and efficient dense SLAM system. Though the above-mentioned methods make great progress in reconstructing static scenes, they are prone to failure when deployed in dynamic environments. We take a step further to address the problem by continually learning what to memorize and what to forget.

\noindent{\bf NeRF construction in changing environments.}
Although there is no prior work specified at tackling the NeRF-based SLAM deployment in dynamic environments, some research has been conducted to train a neural radiance field under changing environments. NeRF in the wild~\cite{martin2021nerf} pioneers the use of appearance embedding to handle environmental changes. The solution is frequently adopted in follow-up works~\cite{turki2022mega, tancik2022block, zhenxing2022switch, suzuki2023federated}. There are also works that target the reconstruction of the spatial-temporal 4D field of the environment~\cite{li2021neural, tretschk2021non, gao2021dynamic, fang2022fast}, where per-point radiance and motion are integrated into the neural representation. Recently, CLNeRF~\cite{cai2023clnerf} leverages continual learning to progressively update the scene representation to adapt to changing environments. Similarly, DynaMoN~\cite{karaoglu2023dynamon} applies motion segmentation to DROID-SLAM~\cite{teed2021droid} to achieve accurate camera pose estimation in dynamic environments. The estimated poses are then utilized for offline optimization of a 4D NeRF. In contrast, our work directly builds the neural radiance field on the fly in dynamic environments and targets a much more challenging SLAM problem. The invariant information within the environment is stored adaptively under different circumstances.

\section{Preliminaries}
\label{sec:pre}
The central idea of this work is a general framework to mitigate the discrepancy between observations and the stored map under changing environments. The framework is expected to distinguish among past observations what are the invariant features to memorize and what are the changing areas to forget. Through this manner, only reliable features that meet the multi-view consistency over a long period will be reserved, and only errors outside the dynamic areas will be back-propagated to update the pose and map parameters. In practice, we resort to a continual learning fashion, where a neural map and a motion status classifier are trained on the fly that distill knowledge from sequential observations into compact networks. The map will serve as a global memory of the scene radiance, whereas the classifier will serve as a dynamic object detector. The iterative optimization of both networks defines the memorization-forgetting loop for a robust neural SLAM system.

\noindent\textbf{Learn to memorize}
Following the recent progress of neural SLAM, we adopt the experience-replay based continual learning for test-time map optimization, where a set of keyframes are stored explicitly for back-propagating errors to the network parameter optimization. The keyframes can be viewed as a compressed knowledge of past experience. The map can then memorize the knowledge through gradient-based optimization given the constantly replayed keyframe buffer.

\noindent\textbf{Learn to forget}
As neural networks exhibit high plasticity, past knowledge can be easily forgotten during constant distribution shift~\cite{Yan2021iccv}. We expect the framework to adaptively forget the scene dynamics while preserving the invariant information. Note that the past knowledge is controlled by the replayed keyframes, we utilize a classifier to identify the areas within stored keyframes that have been changed. The forgetting in the neural mapping naturally undergoes if the dynamic areas on each keyframe are prohibited from replaying. The classifier should be instance-wise and allow efficient adaptation once the environmental changes occur. The updated motion status can then be passed to the stored keyframes to enforce forgetting.

\section{Method}
\label{sec:method}
\begin{figure}[t]
    \centering
    \includegraphics[width=\textwidth]{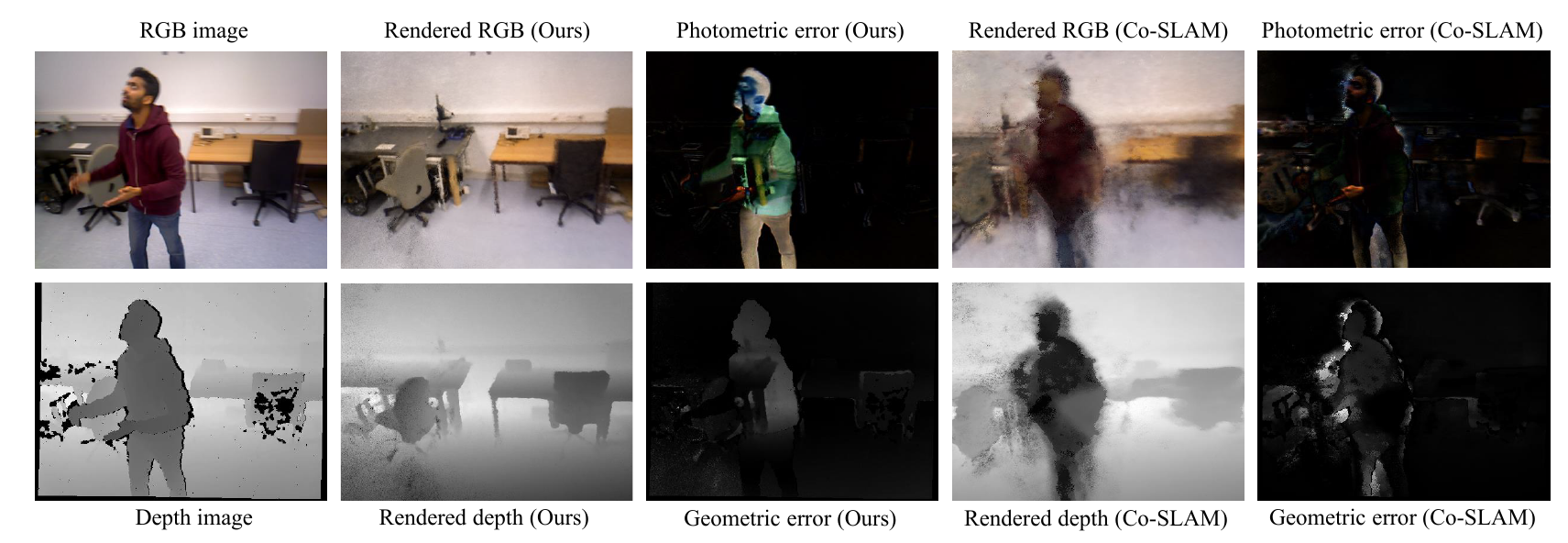}
    \caption{The optimization divergence of either camera pose, neural map, or motion status classifier will lead to high photometric and geometric errors.}
    \label{fig:coslam}
\end{figure}

Fig.~\ref{fig:classifier} shows the overview of our dynamic SLAM framework. In practice, streaming RGB-D images $\{I^t, D^t\}_{t=1}^N$ with known camera intrinsic $K$ are taken as inputs, and a neural radiance field $f(\mathbf{x};\theta^t_M)$ is updated continually to memorize the static part of the environment. The key to our robust SLAM framework in dynamic environments is a continually learned binary classifier $g(\mathbf{z};\theta^t_C)$. The past knowledge to be forgotten will be determined through the classifier by identifying inconsistency induced by object movement.
Note that the optimization of camera pose $\xi^t$, neural map $\theta_M^t$, and motion status classifier $\theta_C^t$ all rely on the discrepancy between rendered and observed RGB-D images. As illustrated in Fig.~\ref{fig:coslam}, the tight coupling of these three variables makes the optimization inherently ambiguous: the divergence of any variables leads to a high discrepancy. In this section, we begin by introducing the photometric and geometric constraints through volume rendering, followed by how these constraints propagate gradients for optimizing the camera pose, map, and classifier parameters iteratively. We argue that this ambiguity can be alleviated with the neural representations of the map and the classifier, where the continuous representations exhibit promising generalization ability and enforce temporally consistent predictions.

\subsection{Formulation}
The objective is to guarantee a photorealistic and multi-view consistent map representation by minimizing the photometric and geometric errors in static areas between the rendered images and observed images as:
\begin{equation}
  L_{pho}=\frac{1}{|H|} \sum_{(u, v) \in H} \left| I \left[u, v \right]-\hat{I}(\xi, \theta_M) \left[u, v \right]\right|,
\label{eq:objective_pho}
\end{equation}
\begin{equation}
  L_{geo}=\frac{1}{|H|} \sum_{(u, v) \in H} \left| D \left[u, v \right]-\hat{D}(\xi, \theta_M) \left[u, v \right]\right|,
\label{eq:objective_geo}
\end{equation}
where $H \subset \mathbbm{1}(R^t)$ is the set of static samples given motion segmentation mask $R^t$ indicated by the classifier. $\hat{I}(\xi^t, \theta^t_M)$ and $\hat{D}(\xi^t, \theta^t_M)$ are predicted color and depth images through volume rendering.

In this paper, we follow Co-SLAM~\cite{wang2023co} to model the scene as a truncated signed distance field \(s\) with color values \(\mathbf{c}\) as $f(\mathbf{x};\theta)\rightarrow (\mathbf{c},s)$. Nevertheless, the proposed method can also adopt the conventional NeRF representation with density-based volume rendering as~\cite{mildenhall2020nerf}. In this work, the learnable map parameters $\theta=\{\alpha, \phi, \tau\}$ denote the encoder $\mathcal{V}(\mathbf{x};\alpha)$ for grid features and the decoders for color and geometry. The volume rendering for color and depth prediction is a weighted sum of all samples $\{\mathbf{p}_i\}_{i=1}^N$ along a ray determined by the pixel coordinate and the camera origin as:
\begin{equation}
  \hat{I}(\xi, \theta_M)[u,v] = \frac{1}{\sum_{i=1}^{N} w_i} \sum_{i=1}^{N} w_i \mathbf{c}_i,
\label{eq:rendering_pho}
\end{equation}
\begin{equation}
  \hat{D}(\xi, \theta_M)[u,v] = \frac{1}{\sum_{i=1}^{N} w_i} \sum_{i=1}^{N} w_i d_i,
\label{eq:rendering_geo}
\end{equation}
where $w_i$ is the computed weight; $d_i$ is the distance between the sampled point $\mathbf{p}_i$ and the corresponding camera origin. It should be noted that in Co-SLAM~\cite{wang2023co}, the weight $w$ is computed by multiplying two Sigmoid functions as:
\begin{equation}
  w = \sigma\left(\frac{s}{\lambda_{tr}}\right) \sigma\left(-\frac{s}{\lambda_{tr}}\right).
\label{eq:weight}
\end{equation}
where \(\lambda_{tr}\) is the truncation distance.

\subsection{Motion-aware Tracking and Mapping}
\label{subsec:ba}
Besides the photometric and geometric constraints in Eq.~\ref{eq:objective_pho} and~\ref{eq:objective_geo}, we also adopt the SDF losses $L_{sdf}$ in near-surface and free space $L_{free}$ along with the feature smoothness loss $L_{smooth}$ in Co-SLAM~\cite{wang2023co} for further constraints.

The SDF loss is employed to enhance the quality of map reconstruction, where the depth values of pixels are utilized as SDF approximations. For points within the truncated region \((|D[u, v] - \hat{D}[u, v]| \leq tr)\): 

\begin{equation}
    L_{sdf} = \frac {1}{|R_H|} \sum _{r\in R_H} \frac {1}{|S_r^{tr}|}\sum _{p \in S_r^{tr}} \big (s_p - (D[u, v] - \hat{D}(\xi, \theta_M)[u, v])\big )^2
\end{equation}
where \( R_H \) represents the set of rays corresponding to the static pixel set \( H \).

For points outside of the truncation region \((|D[u, v] - \hat{D}[u, v]| > tr)\), we calculate the free space loss: 

\begin{equation}
    L_{free} = \frac {1}{|R_H|} \sum _{r\in R_H} \frac {1}{|S_r \setminus S_r^{tr}|}\sum _{p \in S_r \setminus S_r^{tr}} (s_p - tr)^2,
\end{equation}

To mitigate the noisy reconstruction of unvisited areas induced by hash collision, a smoothness loss for the interpolated features is applied:

\begin{equation}
    L_{smooth} = \frac {1}{|\mathcal {G}|}\sum _{p \in \mathcal {G}} \Delta _x^2 + \Delta _y^2 + \Delta _z^2,
\end{equation}
where \(\Delta x, y, z = V_\alpha(p + \epsilon_{x, y, z}) - V_\alpha(p)\) refers to the change in feature metrics between adjacent vertices sampled on the hash-grid.

We apply ADAM optimizer on the weighted sum of these five loss terms:
\begin{equation}
	\label{eq:overall_loss} 
    L = L_{pho}+\lambda_1 L_{geo}+\lambda_2 L_{sdf}+\lambda_3 L_{free}+\lambda_4 L_{smooth}.
\end{equation}

Note that in the loss function, only errors in the static areas will be back-propagated for optimizing the pose and map parameters, where the motion segmentation mask $R^t$ is inferred using the instance-wise classifier $g(\mathbf{z};\theta^t_C)$ in Sec.~\ref{subsec:motion_state}. During the tracking process, we randomly select $|H_t|$ pixels within the static area of the current frame and optimize the estimated camera pose by minimizing the objective function with fixed map and classifier parameters. 

\begin{figure}[t]
    \centering
    \includegraphics[width=\textwidth]{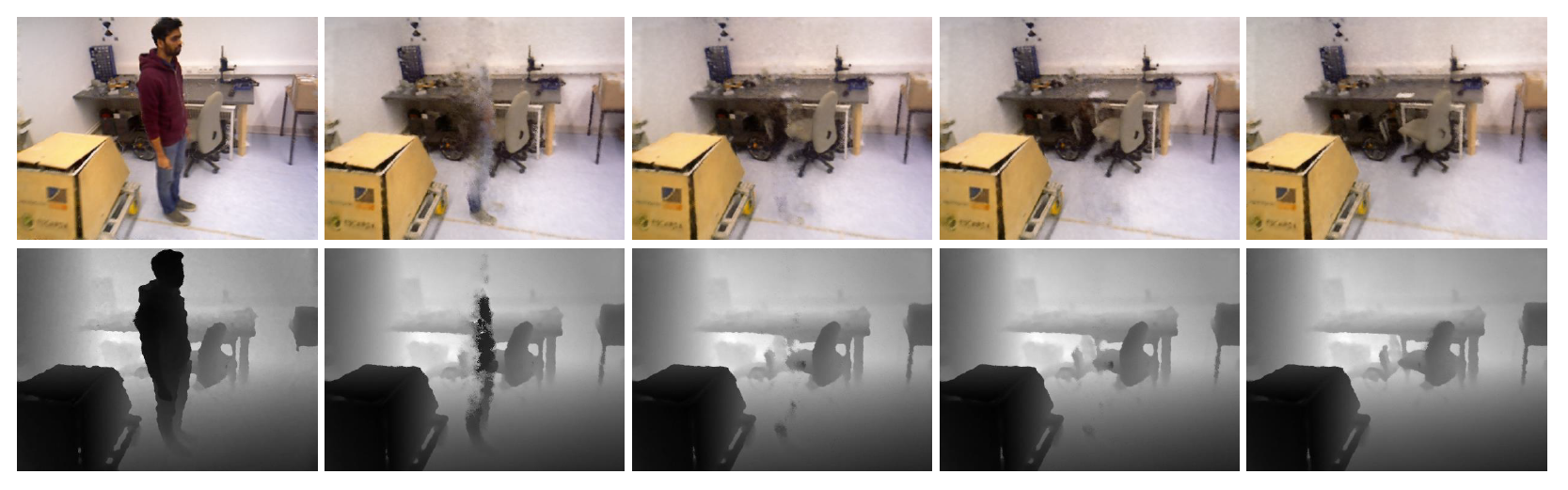}
    \caption{The rendering results given the continually learned map indicate the gradual forgetting of the dynamic object.}
    \label{fig:forgetting}
\end{figure}

Similarly, the bundle adjustment for jointly optimizing camera poses and map parameters is carried out by randomly selecting $|H_b|$ pixels within the static area of the past keyframes and the current frame. Stored keyframes serve as past experience to avoid catastrophic forgetting~\cite{Yan2021iccv,sucar2021imap}. As the error of pixels in the dynamic areas is inhibited from back-propagation, only knowledge in the static areas will be reserved in the neural map. As illustrated in Fig.~\ref{fig:forgetting}, the moving object will be gradually forgotten in the neural map given constantly propagated errors from recent observations.




\subsection{Updating of Object Motion Status}
\label{subsec:motion_state}
As illustrated in Fig.~\ref{fig:classifier}, the motion status is inferred at the instance level. An image $I^t$ is decomposed into $K^t$ segments $\{S_k^t\}_{k=1}^{K^t}$ using FastSAM~\cite{zhao2023fast}, where each segment is then fed to a visual context encoder~\cite{radford2021learning, li2023blip, oquab2023dinov2}. Such decomposition and encoding process turns the image into $K^t$ vectors $\{\mathbf{z}_k^t\}_{k=1}^{K^t}$ to represent $K^t$ different instances. 
Intuitively, we aim to maintain the motion status of each instance across time. Explicitly aggregating information of each instance across views is non-trivial as the dynamic objects lead to inconsistency. We turn to a simple but effective solution that implicitly records the motion status of each instance using a two-layer MLP as $g(\mathbf{z};\theta_C^t)$. The network serves as a classifier that determines the motion status of the corresponding instance feature, where an instance will be treated as a moving object if $g(\mathbf{z}^t_k;\theta_C^t)>0.5$.

\begin{figure}
    \centering
    \includegraphics[width=\textwidth]{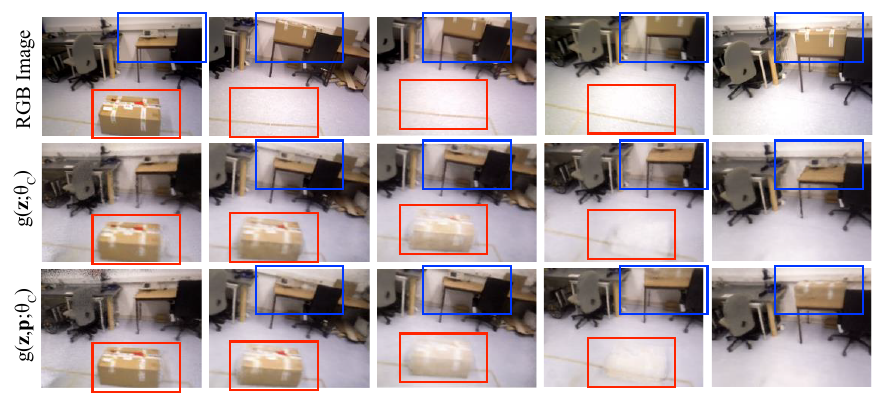}
    \caption{Concatenating the position feature with the semantic embedding allows the classifier to only forget the knowledge regarding the box in the previous position. After being replaced, the box radiance at the latest position will be memorized after a few time intervals.}
    \label{fig:replace}
\end{figure}

The object movement leads to inconsistency between the observation and the stored knowledge. As emitted radiance is view-dependent~\cite{mildenhall2020nerf}, we merely examine the geometry inconsistency between the rendered depth map and the observed one. The moving status $o_k^t$ of the instance $k$ will be treated as \texttt{True} if a certain portion of pixels within the segment $S^t_k$ exhibit large discrepancies as:
\begin{equation}
\label{eq:dynamic_criteria} 
    \frac{\sum_{(u, v) \in S_k^t} \mathbbm{1}\left(\hat{D}(\xi, \theta_M)[u,v] - D[u,v] > t_d\right)}{|S_k^t|} > t_p,
\end{equation}
where $t_d$ and $t_p$ controls the sensitivity of motion segmentation.

The classifier will be updated if the current moving status of any instance $o_k^t$ contradicts the classification result $\mathbbm{1}(g(\mathbf{z}^{t}_k;\theta_C^{t-1})>0.5)$. The network is optimized using the binary cross-entropy loss as:
\begin{equation}
\label{eq:masknet_updating} 
   -\sum_{k=1}^{K^t} \big[o_k^t\log(g(\mathbf{z}^{t}_k;\theta_C^*))+(1-o_k^t)\log(1-g(\mathbf{z}^{t}_k;\theta_C^*))\big].
\end{equation}

Once the classifier is updated, the motion segmentation mask of all keyframes will be updated. The observations regarding the moving object will no longer contribute to the pose and map optimization. As the neural map is continually trained under constant distribution shift, the knowledge regarding the moving object will be gradually forgotten without manual operations.

\noindent{\bf Continual learning of the classifier.}
Catastrophic forgetting is not only an issue that neural SLAM system need to address, it also has a negative impact on the classifier. If we train the classifier using only the objects that appear in the current frame, catastrophic forgetting could occur, causing the classifier to forget previously learned dynamic objects. Therefore, we also maintain a replay buffer for the classifier to avoid this issue. For all instances in the current frame, if an instance is marked as dynamic after passing through a check for geometric inconsistency, it will be added to this replay buffer. When updating the classifier, training is conducted not only with instances from the current frame but also by randomly selecting \( n_c \) instances from the replay buffer as training data. \( n_c \) is typically set to 5 in our experiments.

\begin{figure}[t]
    \centering
    \includegraphics[width=\textwidth]{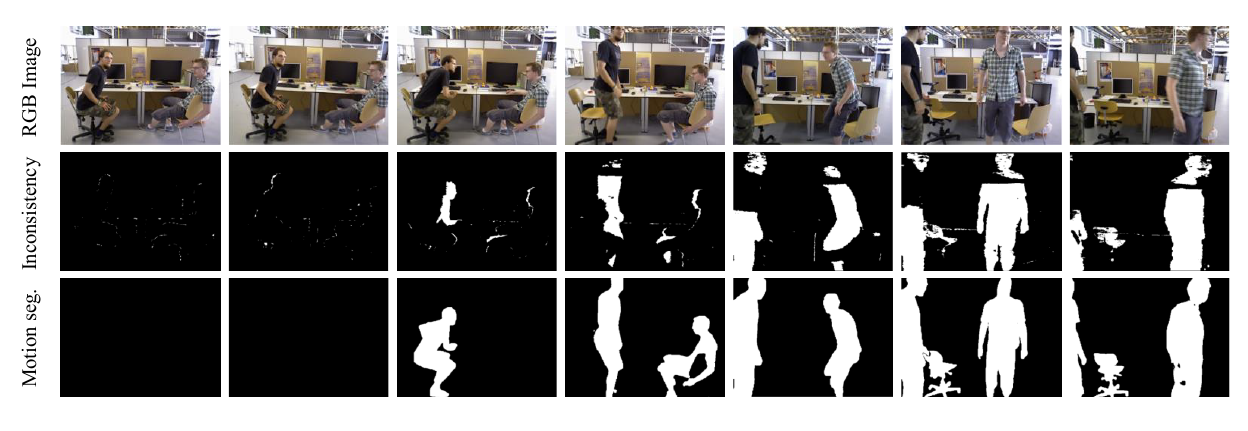}
    \caption{The continual learning of the classifier on the fr3\_walking\_static sequence of TUM RGB-D dataset. A segment is treated as a dynamic part if a certain portion within the segment exhibits high discrepancy.}
    \label{fig:masknet}
\end{figure}

\noindent{\bf Bidirectional check.} Eq.~\ref{eq:dynamic_criteria} identifies a moving object if it is observed in front of the rendered place. However, if an object is removed from the scene or if the object moves away from the camera since the first frame, the observation may fall behind the rendered one. Therefore, we apply a bidirectional check. Similar to Eq.~\ref{eq:dynamic_criteria}, we decompose the rendered image $\hat{D}(\xi, \theta_M)$ into $\hat{K}^t$ segments $\{\hat{S}_k^t\}_{k=1}^{\hat{K}^t}$ and apply the consistency check as:
\begin{equation}
\label{eq:dynamic_criteria_inverse} 
    \frac{\sum_{(u, v) \in \hat{S}_k^t} \mathbbm{1}\left(D[u,v] - \hat{D}(\xi, \theta_M)[u,v] > t_d\right)}{|\hat{S}_k^t|} > t_p.
\end{equation}

\noindent{\bf Incorporation of prior knowledge.}
One promising characteristic of the open-set classifier is that we can apply prior knowledge through pre-training to pre-define the instance motion status while allowing adaptation to the current state. Specifically, we can train a classifier $\theta_C^{pre}$ on specific categories, e.g., human, car, animal. The object motion status can then be identified as "dynamic" if either the pre-trained one or the online-learned one agrees.

\noindent{\bf Maintaining the instant object state.}
In the above-mentioned setting, we would label any object that has been moved as "dynamic" even if it is replaced in a new area and remains static afterward. It is due to the fact that the classifier merely establishes the mapping between the instance feature and the corresponding motion state. Thanks to the flexible structure of the classifier, we can concatenate an additional position feature along with the semantic embedding as the inputs. The classifier turns to record the motion state of an object at a specific position (object center $\mathbf{p}_k^t$ for instance) as $g(\mathbf{z}_k^t, \mathbf{p}_k^t; \theta_C^t)$. As we show in the experiment, this simple strategy leads to a neural map that forgets the removed object and memorizes the object at the new position gradually.

\section{Experiments}
\label{sec:experiments}
We evaluate the proposed method on multiple sequences of TUM RGB-D dataset~\cite{sturm12iros} and Bonn RGB-D dataset~\cite{palazzolo2019iros}. The quality of estimated poses, neural map, and motion status reasoning are analyzed qualitatively and quantitatively. As the first NeRF-SLAM-based method that tackles challenging dynamic environments, we compare the proposed method against traditional methods of ReFusion~\cite{palazzolo2019refusion}, StaticFusion~\cite{scona2018staticfusion}, and DynaSLAM~\cite{bescos2018dynaslam} that are designed specifically to dynamic environments. We also compare against our codebase of Co-SLAM~\cite{wang2023co} to see how the proposed strategies guarantee robust camera tracking with promising reconstruction quality.

\subsection{Experimental Setup}
The experiments are conducted on a desktop PC with an Intel i9-12900K CPU, an NVIDIA RTX 3090 GPU, and 64GB memory. The keyframe is automatically stored every 5 frames. 

Specific to implementation details, the number of sampling points along each ray $N=128$, sampling pixels during tracking and bundle adjustment \(H_t=1024, H_b=2048\), truncation distance \(tr=10cm\), threshold used for determining moving status \(t_d=0.3, t_p=0.05\). For the weights of each loss in Eq.~\ref{eq:overall_loss}, we follow the settings of Co-SLAM as: \(\lambda_1=0.1, \lambda_2=5000, \lambda_3=10, \lambda_4=1e-8\).

\begin{table}[t]
\centering
\caption{Comparisons of ATE (RMS) against traditional SLAM algorithms designed for deployment in dynamic environments.}
\label{tab:tracking}
\resizebox{\textwidth}{32 mm}{

\begin{tblr}{
  column{3} = {c},
  column{4} = {c},
  column{5} = {c},
  column{6} = {c},
  column{7} = {c},
  cell{2}{1} = {r=9}{},
  cell{11}{1} = {r=6}{},
  hline{1-2} = {-}{},
  hline{11,17} = {-}{},
}
    & \textbf{Sequence}    & \textbf{ReFusion}~\cite{palazzolo2019refusion} & \textbf{StaticFusion}~\cite{scona2018staticfusion} & \textbf{DynaSLAM}~\cite{bescos2018dynaslam} & \textbf{Co-SLAM}~\cite{wang2023co} & \textbf{Ours}  \\
\begin{sideways}Bonn\end{sideways} & balloon              & \uline{0.175}              & 0.233                  & \textbf{0.050}     &      0.308            & 0.206  \\
    & balloon2             & 0.254              & 0.293                  & \uline{0.142}      &    0.290              & \textbf{0.136} \\
    & kidnapping\_box      & 0.148              & 0.336                  & \textbf{0.026}     &       \uline{0.095}           & 0.112  \\
    & kidnapping\_box2     & 0.161              & 0.263                  & \textbf{0.033}     &     0.118             & \uline{0.104}  \\
    & crowd                & \uline{0.204}      & 3.586                  & 1.065              &    fail              & \textbf{0.116} \\
    & crowd2               & \textbf{0.155}     & 0.215                  & 1.217              &    fail              & \uline{0.200}  \\
    & crowd3               & \uline{0.137}     & 0.168                  & 0.835              &    fail              & \textbf{0.107}  \\
    & person\_tracking     & \uline{0.289}      & 0.484                  & 0.714              &    fail              & \textbf{0.274} \\
    & synchronous          & \uline{0.441}      & 0.446                  & 0.977              &    0.634              & \textbf{0.130} \\
\begin{sideways}TUM\end{sideways}  & fr3\_walking\_static & 0.017              & \uline{0.015}          & \textbf{0.014}     &        fail          & 0.025          \\
    & fr3\_walking\_xyz    & 0.099              & 0.093          & \uline{0.085}     &    fail              & \textbf{0.076}          \\
    & fr3\_walking\_halfsphere  & 0.104  &  0.681 &  \uline{0.084}  &  fail  &  \textbf{0.079} \\
    & fr3\_sitting\_static & \uline{0.009}              & 0.014                  & \uline{0.009}      &    0.011              & \textbf{0.007} \\
    & fr3\_sitting\_xyz    & 0.040              & 0.039                  & \textbf{0.009}     &    0.020              & \uline{0.018}  \\
    & fr3\_sitting\_halfsphere  & 0.110  &  0.041 &  \textbf{0.017}  & 0.042  &  \uline{0.039}
\end{tblr}}
\end{table}

\begin{figure}[t]
    \centering
    \includegraphics[width=\textwidth]{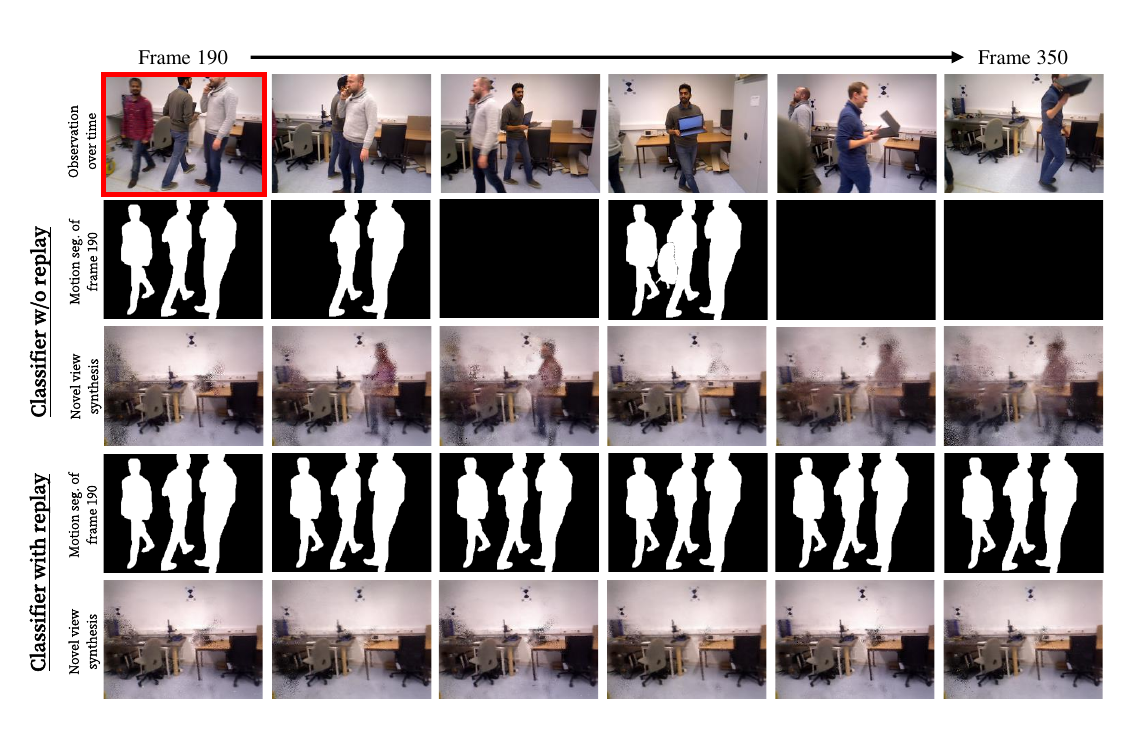}
    \caption{The comparison regarding whether the classifier undergoes continual learning. The 2nd and 4th rows illustrate the classifier's determination results for dynamic instances in frame 190 at different times. The 3rd and 5th rows display the rendering results to demonstrate changes in the map.}
    \label{fig:classifier_replay}
\end{figure}

\subsection{Tracking and Mapping}
As illustrated in Fig.~\ref{fig:coslam}, the photometric and geometric errors in dynamic areas lead to inherent ambiguities as they will be back-propagated to modify the pose estimates and mapping results. As tracking and mapping are coupled during the optimization, the erroneous tracking and the fusion of dynamic objects into the map would eventually lead to system collapse. The proposed method, on the other hand, achieves robust camera tracking results even in the challenging 'crowd' sequences. As presented in Tab.~\ref{tab:tracking}, the proposed method achieves better results compared against the feature-based DynaSLAM~\cite{bescos2018dynaslam} and the dense SLAM systems of ReFusion~\cite{palazzolo2019refusion} and StaticFusion~\cite{scona2018staticfusion}. The effects arising from the dynamic objects are well alleviated by the proposed motion status classifier.

\noindent\textbf{The forgetting and memorization.} 
The forgetting mechanism is demonstrated in Fig.~\ref{fig:forgetting}. Once the classifier identifies the man as a moving object, the knowledge will no longer act on the map parameters. Meanwhile, the areas occluded by the man will appear once the man goes away, and the inconsistency between the rendered image and the observation in other views will contribute to the map updating. As demonstrated in Sec.~\ref{subsec:motion_state}, one interesting fact is that we can also maintain the updated object status through a simple coordinate concatenation. As illustrated in Fig.~\ref{fig:replace}, the box will be forgotten after being kidnapped, and reappear on the map after being placed.

\begin{figure}[t]
    \centering
    \includegraphics[width=\textwidth]{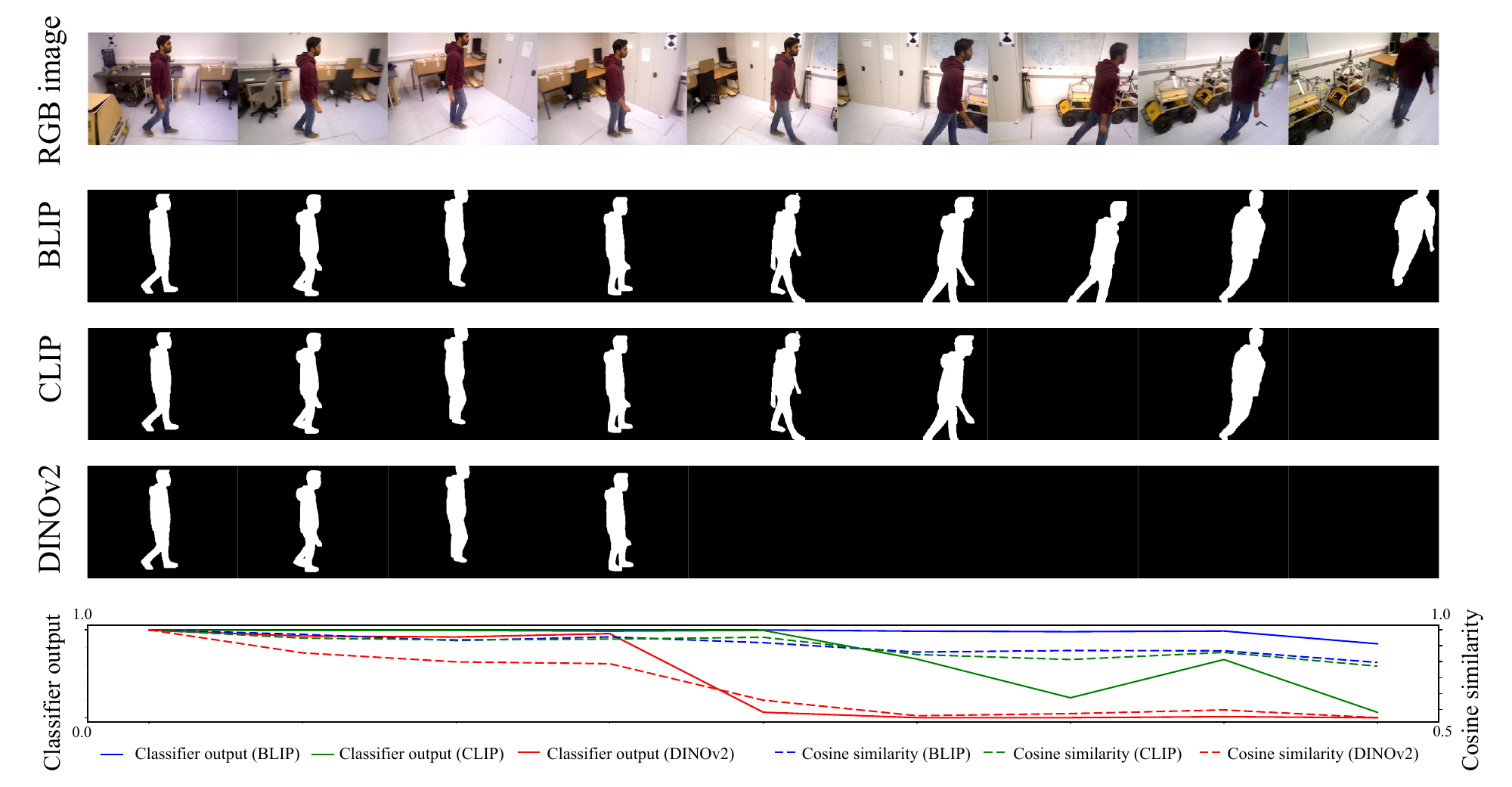}
    \caption{The different choices of visual encoder lead to diverse behaviors for predicting object motion status in incoming frames.}
    \label{fig:ablation_encoder}
\end{figure}

\subsection{Motion Segmentation}
\label{subsec:motion_segmentation}
As illustrated in Fig.~\ref{fig:masknet}, the continual learning of the classifier leads to dynamic updating of the object motion status. The successive changes in motion status can be well reflected by the geometric inconsistency, and the classifier can quickly identify the motion changes and learn the instant motion status. The encoded feature of masked objects can well distinguish between two people even if they fall into the same "person" category.

\noindent \textbf{Continual learning of the classifier.} 
As mentioned in Sec.~\ref{subsec:motion_state}, the classifier could also encounter catastrophic forgetting. Therefore, we implement continual learning for the classifier by maintaining a replay buffer. As demonstrated in Fig.~\ref{fig:classifier_replay}, we select a specific frame (frame 190) for observation. It is evident that the classifier with replay maintains its ability for accurate motion segmentation on the earlier frame over time. In contrast, the classifier without continual learning is likely to perform poorly on the previous training data after updates, segmenting the dynamic instances incorrectly. Specifically, the person in red gradually walks out of the frame (see the second column of Fig.~\ref{fig:classifier_replay}). Meanwhile, the classifier undergoes an update. Without replay, training solely on instances from the current frame, the classifier forgets that the person in red is also dynamic. This oversight leads to the person erroneously appearing in the map, further affecting subsequent calculations of geometric inconsistency. This experiment strongly proves that the continual learning strategy designed for the classifier is both effective and necessary.

\noindent\textbf{The choices of visual encoders.} To further understand how the motion segmentation behavior is affected by encoded features, we look into the motion prediction given a fixed classifier. As illustrated in Fig.~\ref{fig:ablation_encoder}, after training the classifier given supervisory signal from the frame $I^{40}$, we freeze the network $\theta_C^{40}$ and predict the motion status of the human with the following observations $I^{40},\cdots, I^{280}$ every 30 frames as $g(\mathbf{z}_k^{40:280};\theta_C^{40})$. We compare three different visual encoders (BLIP~\cite{li2023blip}, CLIP~\cite{radford2021learning}, and DinoV2~\cite{oquab2023dinov2}) and visualize the predicted motion state and the encoded feature similarity compared to the frame $I^{40}$. Apparently, the powerful semantic information of language-aligned visual context presents promising temporal consistency even if the object poses change drastically. The consistent feature embedding across views makes the implicit classifier instance-aware and easy to adapt. We argue that the promising prediction capability leads to robust camera tracking under complex scene dynamics.

\begin{figure}[t]
    \centering
    \includegraphics[width=\textwidth]{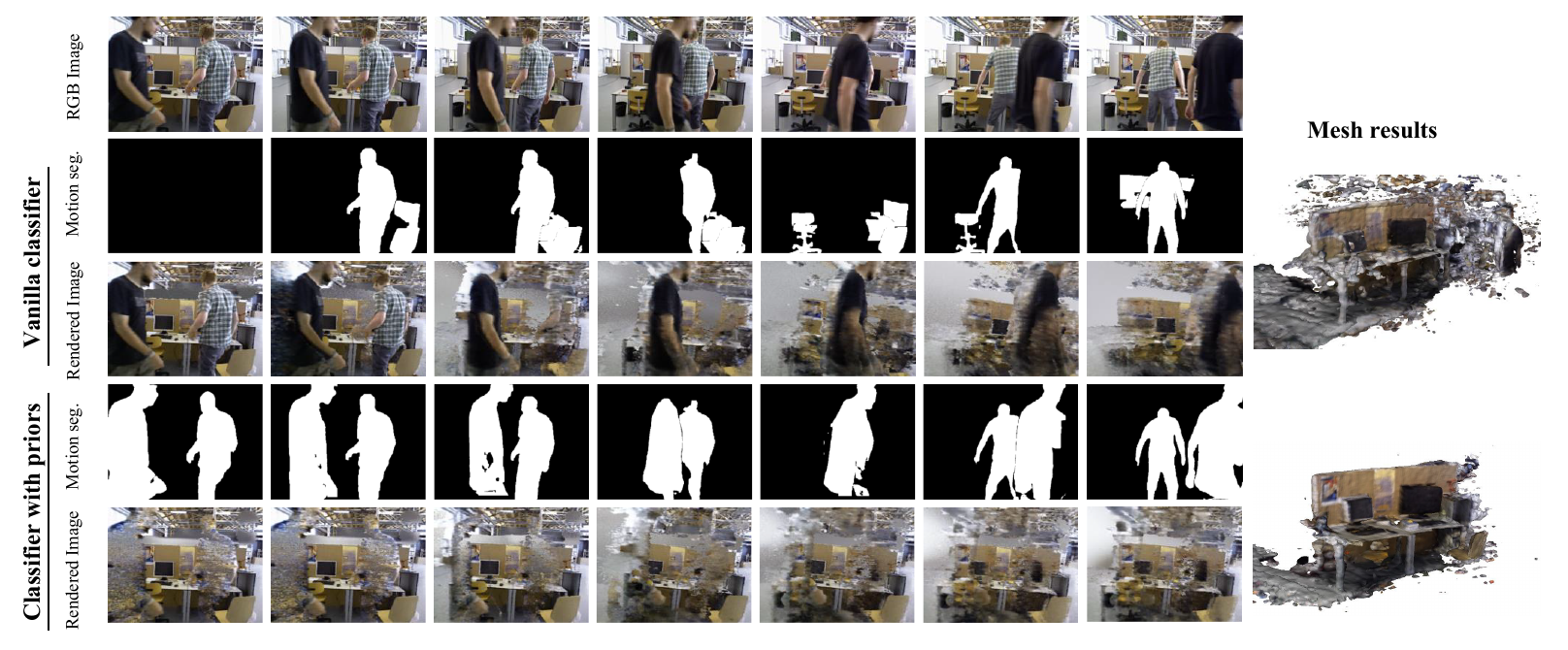}
    \caption{The incorporation of prior knowledge leads to more robust camera tracking under high dynamic environments.}
    \label{fig:prior}
\end{figure}

\noindent{\bf Incorporating pre-defined prior knowledge.} 
As demonstrated in Sec.~\ref{subsec:motion_state}, the proposed framework allows convenient incorporation of prior knowledge by pre-training a classifier on specific categories. An exemplary case is illustrated in Fig.~\ref{fig:prior}. Here we experiment using a segment after frame 410 from the freiburg3 walking static sequence, where individuals occupy a significant portion of the frame from the outset. By pre-defining humans as dynamic objects and pre-training a classifier using a human matting dataset~\cite{chen2018semantic}, the ambiguity induced by human motion can be alleviated at the very beginning. On the contrary, the classifier without prior knowledge will have trouble figuring out the actual factor that leads to inconsistency in such a challenging case. As illustrated in the figure, the accurate static map can be reconstructed even though the moving objects occupy nearly half of the frame when the stream begins.

\section{Conclusion}
In this paper, we address the dynamic SLAM problem with a neural map representation. By learning an instance-aware classifier online that implicitly records the per-object motion status, the invariant information within the observations can be continually distilled to the neural map, where the interference induced by dynamic objects is best alleviated. The iterative optimization of camera pose, map, and classifier parameters forms a robust SLAM framework in challenging dynamic environments.

\clearpage

\section*{Acknowledgement}
We gratefully acknowledge the anonymous reviewers and AC for their valuable comments and suggestions. This work is supported by NSFC (U22A2061, 62176010) and 230601GP0004.

%
%
\bibliographystyle{splncs04}
\bibliography{main}

\newpage
\setcounter{page}{1}

\title{Learn to Memorize and to Forget: A Continual Learning Perspective of Dynamic SLAM \large Supplementary Material} 

\titlerunning{Learn to Memorize and to Forget}

\author{Baicheng Li\inst{1} \and
Zike Yan\inst{2}$^\dag$ \and
Dong Wu\inst{1} \and
Hanqing Jiang\inst{3} \and
Hongbin Zha\inst{1}$^\dag$ }

\authorrunning{B. Li et al.}

\institute{National Key Lab of GAI, School of IST \\
PKU-SenseTime Machine Vision Joint Lab \\
Peking University\and
AIR, Tsinghua University \and
SenseTime Research \\
}

\maketitle

\section{Supplementary Results}

\subsection{Segmentation Pruning}

Segmentation with FastSAM~\cite{zhao2023fast} leads to varying granularities. A person may be separated into different parts as arms, legs, body, and head. In such a case, we wish to retain the instance-level segmentation instead of the part-level decomposition, thereby reducing the number of classifier updates.

\begin{figure}[h]
    \centering
    \includegraphics[width=\textwidth]{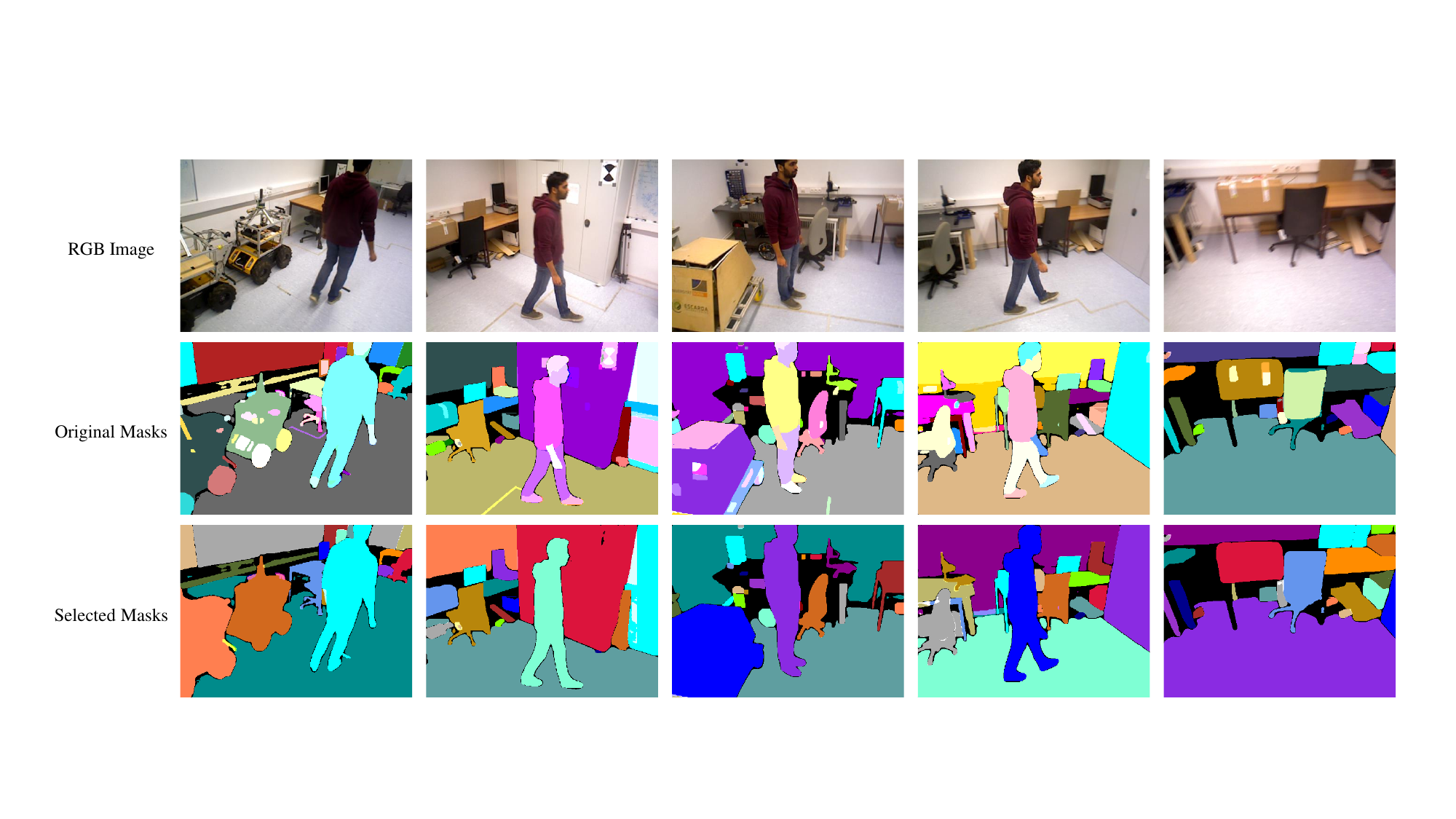}
    \caption{The mask selection strategy helps to reduce a significant number of unnecessary masks, thereby lessening interference to the system.}
    \label{fig:masknet_selection}
\end{figure}

Specifically, as illustrated in Fig.~\ref{fig:masknet_selection}, for any two segments \( R_1 \) and \( R_2 \), we consider \( R_1 \) to be a part of \( R_2 \) and delete it if the portion of overlapped areas between \( R_1 \) and \( R_2 \) is larger than \( T_R \) as: 
\begin{equation}
    S_{R_1 \cap R_ 2} > T_R \times S_{R_1}
\end{equation}
where \( T_R \) is set to 0.9 in our experiments.

\subsection{Mesh Evaluation}
\begin{figure}[h]
    \centering
    \includegraphics[width=\textwidth]{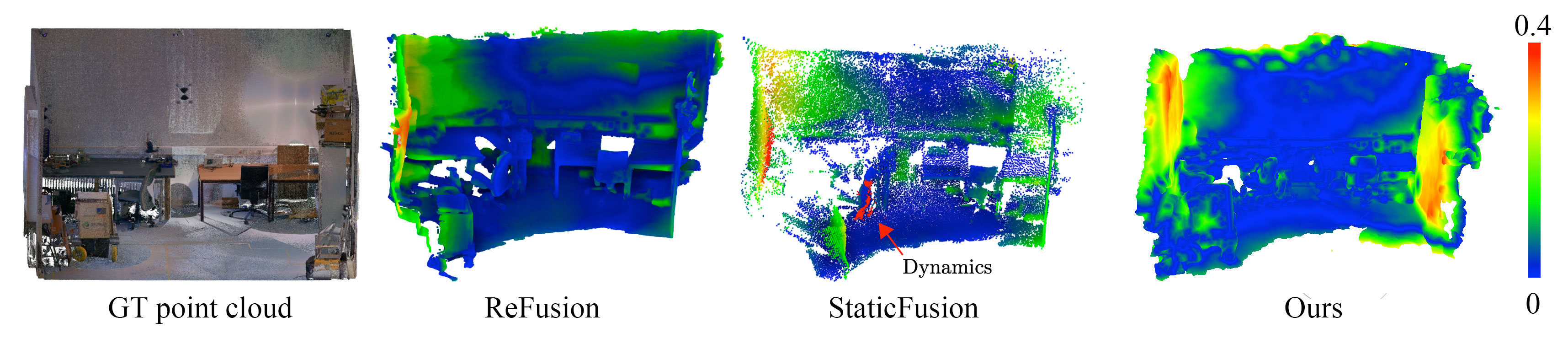}
    \caption{Additional comparison results of mapping and tracking.}
    \label{fig:mesh_error_map}
\end{figure}

\begin{figure}[h]
    \centering
    \includegraphics[width=\textwidth]{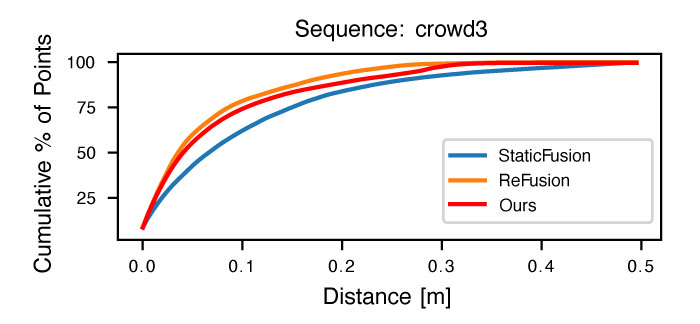}
    \caption{Additional comparison results of mapping and tracking.}
    \label{fig:mesh_evaluation}
\end{figure}

We follow the dense dynamic SLAM system of ReFusion~\cite{palazzolo2019refusion} to evaluate the mesh results quantitatively. As shown in Fig.~\ref{fig:mesh_error_map} and ~\ref{fig:mesh_evaluation}, our method achieves comparable results with ReFusion~\cite{palazzolo2019refusion} and outperforms StaticFusion~\cite{scona2018staticfusion}. Note that NeRF-based methods mainly focus on realistic rendering (as presented in our main paper and supp.~video) instead of surface reconstruction, the quantitative results sufficiently verify our map quality in dynamic environments.

\subsection{Ablation Study on the Replay Training of the Classifier}
\begin{figure}[h]
    \begin{minipage}{0.5\linewidth}
        \centering
        \includegraphics[width=2.5in]{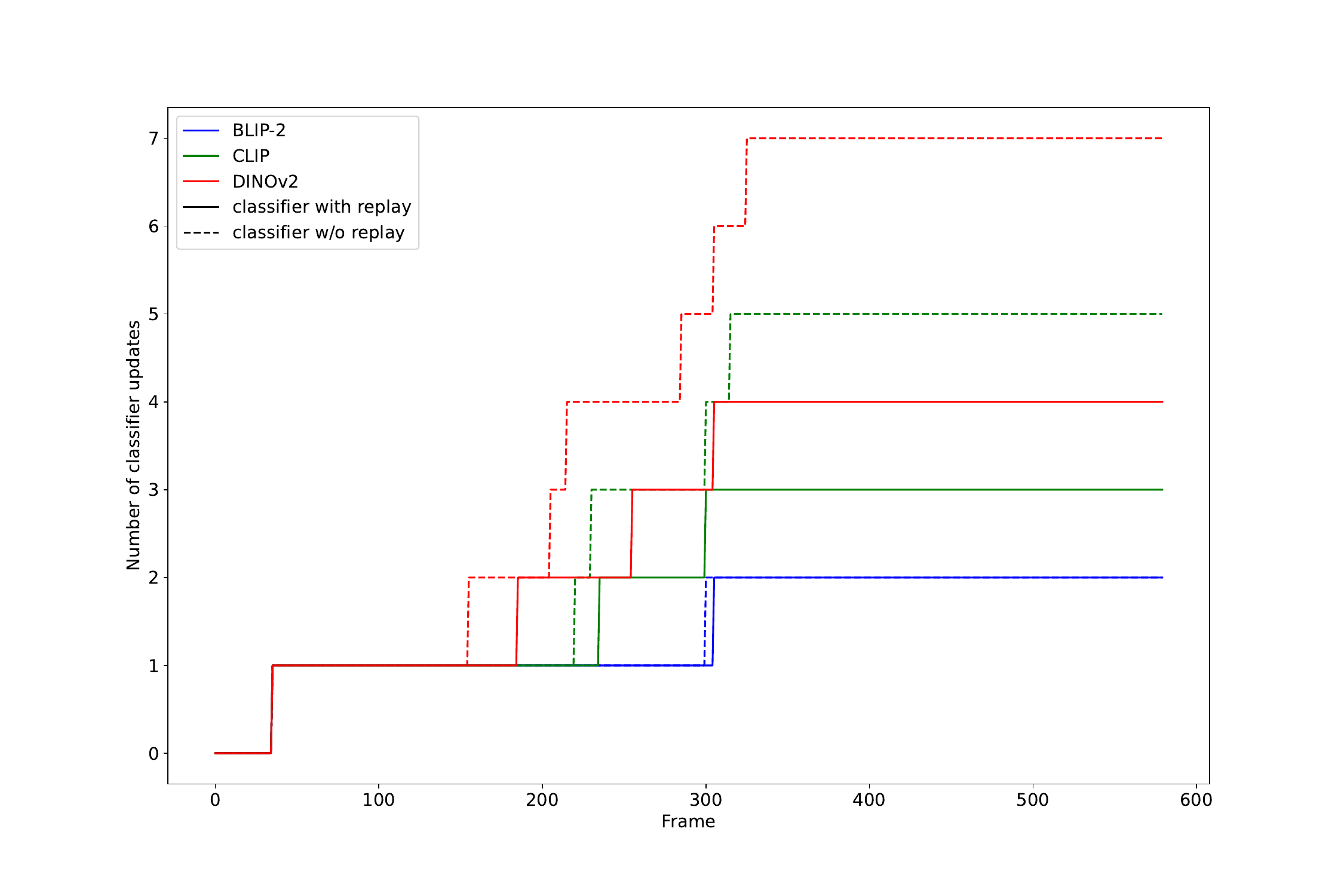}
    \end{minipage}
    \begin{minipage}{0.5\linewidth}
        \centering
        \includegraphics[width=2.5in]{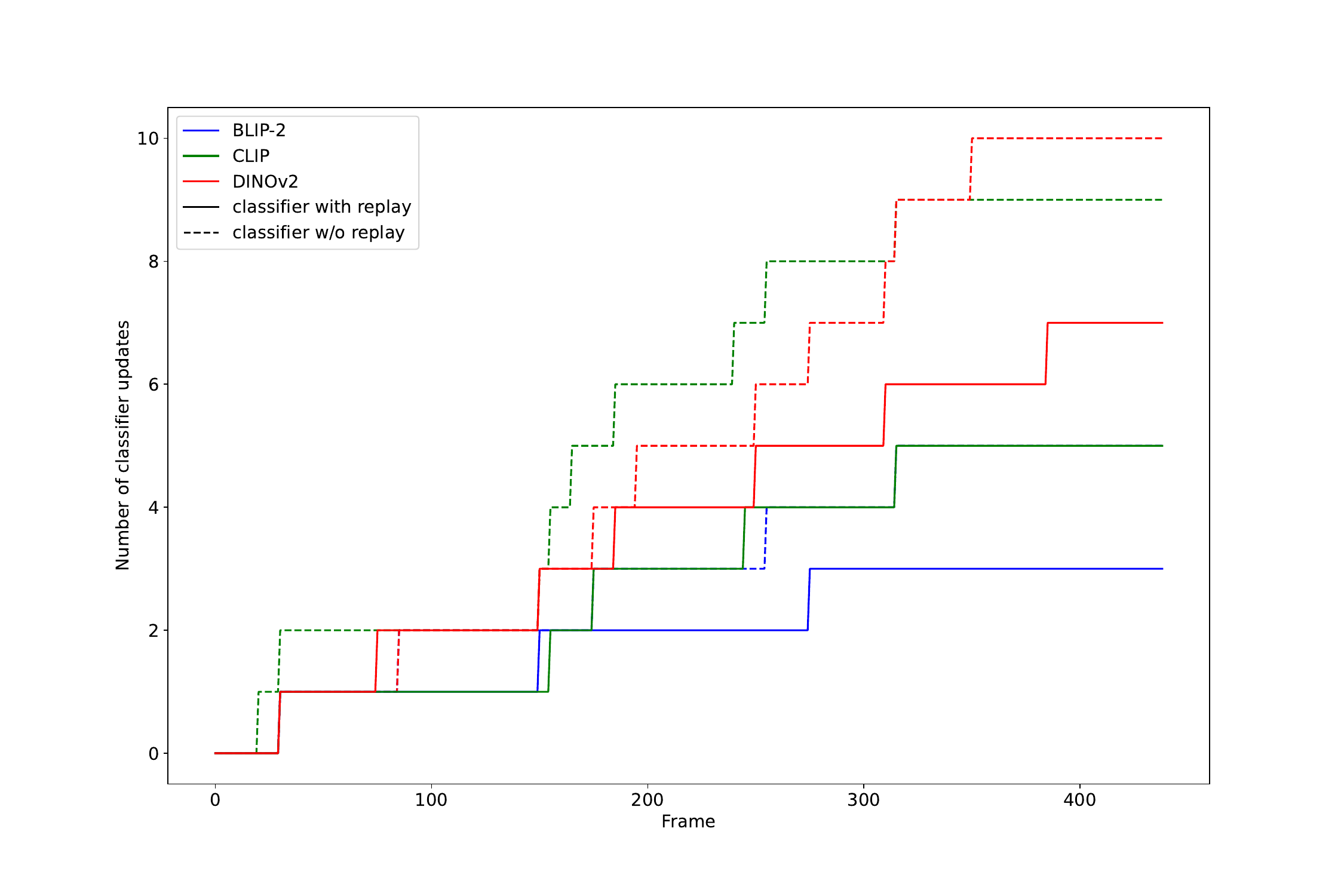}
    \end{minipage}
    \caption{Classifier update times: with and without replay training.}
    \label{fig:update_times}
\end{figure}

We compare the number of updates for the classifier based on whether replay training is performed. The experiment is conducted with different image encoders across two high-dynamic sequences: 'person\_tracking' and 'balloon'. The results are illustrated in Fig.~\ref{fig:update_times}. It is clearly observable that the classifier undergoing replay training requires fewer updates compared to the one without replay, regardless of the image encoder used. Detailed analysis can be found in the "continual learning of the classifier" parts of Sec.~\ref{subsec:motion_state} and Sec.~\ref{subsec:motion_segmentation}.

\subsection{Run-time}

We also tested the average run-time of each component of the system in the two sequences mentioned above. In high-dynamic environments, we effectively balanced the system's speed and accuracy. Our method can run at a frame rate of around 1 fps. In contrast, the frame rates of Co-SLAM~\cite{wang2023co}, iMap~\cite{sucar2021imap}, and NICE-SLAM~\cite{zhu2022nice} are approximately 3 fps, 2 fps, and less than 0.1 fps, respectively.

\begin{table}[ht]
  \centering
    \caption{Average run-time of Bonn dataset.}
  \label{tab:my_label}
  \begin{tabularx}{0.98\textwidth}{XXXX}
    \hline
    Instance feature extraction & Tracking & Bundle adjustment & Classifier updating \\
    \hline
    231ms & 149ms & 161ms & 397ms \\
    \hline
  \end{tabularx}
\end{table}

\subsection{Visualization of Tracking Results}
\begin{figure}[h]
    \centering
    \includegraphics[width=\textwidth]{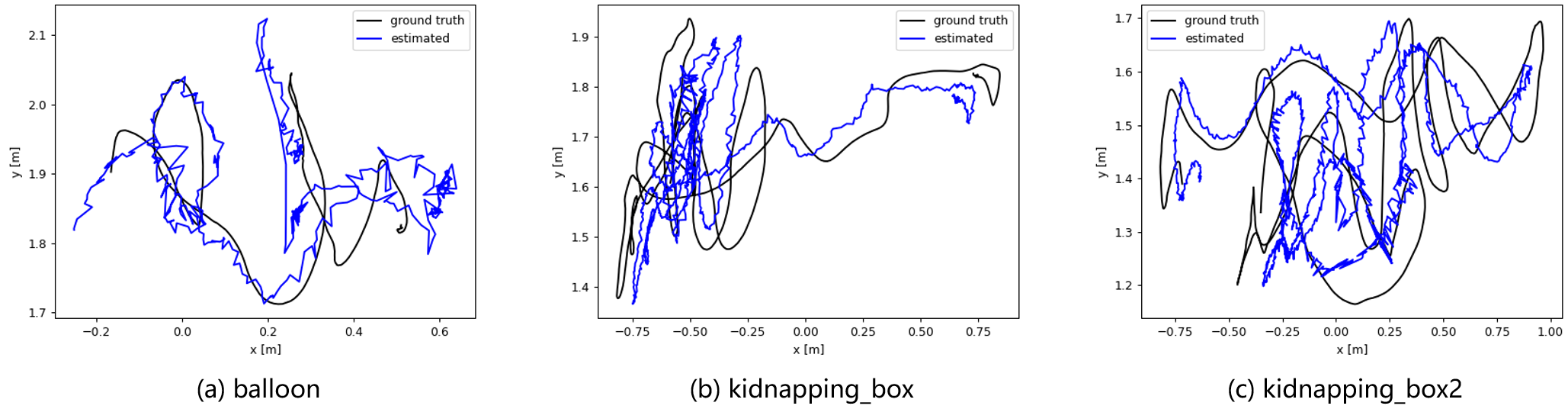}
    \caption{Trajectory plots of three sequences in Bonn RGB-D dataset.}
    \label{fig:trajectory_plot}
\end{figure}

In Table~\ref{tab:tracking} of the main paper, we present the quantitative results of camera tracking, showing that our method achieves higher accuracy compared to other dynamic SLAM approaches. Here, we also provide a qualitative demonstration of these results. Fig.~\ref{fig:trajectory_plot} shows a comparison of our trajectories with the ground truth on three sequences.

\begin{table}[h]
	\begin{minipage}{0.51\linewidth}
		\centering
		
		\caption{ATE (RMS) with different visual encoders.}
		\label{table:encoder_ablation}
		\resizebox{1\textwidth}{!}{
			\begin{tabular}{lccc}
				\hline
				Sequence & BLIP-2 & CLIP & DINOv2 \\ \hline 
				balloon		& 0.206 & 0.211 & 0.228 \\ 
				synchronous	& 0.130 & 0.139 & 0.146 \\
                person\_tracking	& 0.274 & 0.271 & 0.278 \\ \hline
			\end{tabular}
		}
		
	\end{minipage}
	\hfill
	\begin{minipage}{0.475\linewidth}  
		\centering
		
		\caption{ATE (RMS) with and without prior knowledge.
		}
		\label{table:prior_ablation}
		\resizebox{1\textwidth}{!}{
			\begin{tabular}{lcc}
				\hline
				Sequence & w/o prior & with prior \\ \hline 
				balloon		& 0.206 & 0.212 \\ 
				synchronous	& 0.130 & 0.134 \\ 
                person\_tracking	& 0.274 & 0.259 \\ \hline
			\end{tabular}
		}
		
	\end{minipage}
\end{table}

\subsection{Ablation Study on the Visual Encoders and Prior Knowledge}
We tested the impact of different experimental settings on the accuracy of camera tracking. Table~\ref{table:encoder_ablation} and Table~\ref{table:prior_ablation} present the ablation study results on Bonn RGB-D dataset. As shown in Table~\ref{table:encoder_ablation}, using different visual encoders causes slight variations in the tracking results, with BLIP-2~\cite{li2023blip} performing the best and DINOv2~\cite{oquab2023dinov2} performing the worst. Table~\ref{table:prior_ablation} reflects that the presence of prior knowledge has minimal impact on tracking accuracy. In most cases, our method achieves precise camera tracking without prior knowledge. Only in handling extremely challenging scenarios (such as Fig.~\ref{fig:prior} in the main paper), do we need to incorporate prior knowledge.

\section{The Forgetting Issue of Implicit Neural Representations}
The global representation of iMap~\cite{sucar2021imap} results in severe catastrophic forgetting if keyframe-based replay is not deployed, as the distribution shift constantly occurs during the sequential data capturing. On the other hand, subsequent NeRF-based SLAMs like Co-SLAM~\cite{wang2023co} and Point-SLAM~\cite{sandstrom2023point} introduce the local neural representations. They store local features of the scene on grids or points, which to some extent alleviates this negative impact. However, they also employ a global decoder to interpret local features, leading to catastrophic forgetting of the global decoder if keyframe replay is not performed, thereby affecting the operation of the entire system. Therefore, the keyframe buffer lays the foundation of the neural SLAM systems for distilling all past knowledge jointly. We control the replayed buffer through a continually learned classifier so effects from dynamic objects over the past observations can be instantly eliminated, thereby leading to the forgetting of these dynamic objects and maintaining an invariant map to be updated. The methodology is applicable not only to global neural representations but also to representations with discretely-stored local features.

\end{document}